# Frost Prediction Using Machine Learning Methods in Fars Province

Milad Barooni
Department of Computer Science, Engineering and Information Technology
Shiraz University
Shiraz, Iran
m.barooni@cse.shirazu.ac.ir

Koorush Ziarati
Department of Computer Science, Engineering and Information Technology
Shiraz University
Shiraz, Iran
ziarati@shirazu.ac.ir

Ali Barooni
Department of Computer Science, Engineering and Information Technology
Shiraz University
Shiraz, Iran
abarooni@shirazu.ac.ir

*Abstract*—One of the common hazards and issues in meteorology and agriculture is the problem of frost, chilling or freezing. This event occurs when the minimum ambient temperature falls below a certain value. This phenomenon causes a lot of damage to the country, especially Fars province. Solving this problem requires that, in addition to predicting the minimum temperature, we can provide enough time to implement the necessary measures. Empirical methods have been provided by the Food and Agriculture Organization (FAO), which can predict the minimum temperature, but not in time. In addition to this, we can use machine learning methods to model the minimum temperature. In this study, we have used three methods Gated Recurrent Unit (GRU), Temporal Convolutional Network (TCN) as deep learning methods, and Gradient Boosting (XGBoost). A customized loss function designed for methods based on deep learning, which can be effective in reducing prediction errors. With methods based on deep learning models, not only do we observe a reduction in RMSE error compared to empirical methods but also have more time to predict minimum temperature. Thus, we can model the minimum temperature for the next 24 hours by having the current 24 hours. With the gradient boosting model (XGBoost) we can keep the prediction time as deep learning and RMSE error reduced. Finally, we experimentally concluded that machine learning methods will work better than empirical methods and XGBoost model can have better performance in this problem among other implemented.

## I. Introduction

When the temperature reaches zero degree Celsius, due to the possibility of ice nucleation, the possibility of rapture and disintegration in plant cells increases, and sensitive agricultural products can be damaged, which has significant effects on production [1] This phenomenon is called Frost. An example of its occurrence can be seen in Fig 1.

Surveys show that among the 12 strategic products covered by insurance, the most features were related to frost, which was an average of 34.8% of the total compensation. The figures of compensation paid for frost damage in recent years have varied between 613 and 1591 billion Rials. Obviously, these figures do not represent the real damage because firstly, not all farmers and gardeners insure their products, and secondly, insurance funds pay only a part of the damage caused to the products, and the real damage in the country is estimated several times the above figure [2]. Every year, the frost causes a lot of damage in Iran, and Fars province and garden products are one of the most affected.

TABLE I. Different type of frosts [3]

| Frost Type | Characteristics |
|---|---|
| *Radiation* | Clear sky; calm or very little wind; temperature inversion; low dew-pint; air temperature greater than 0° during day. |
| *Advection* | Windy and cloudy; No temperature inversion; low humidity; air temperature can be less than 0° during day |

The dew-pint temperature is the temperature reached when air is cooled until reaches 100 percent relative humidity, and it is a direct measure of the water vapor content of the air. There are two types of frost occurs. Radiation frosts are common occurrences. Due to receiving energy during the day and releasing it at night by the earth, in some special conditions during the night until sunrise, the heat is quickly lost and the heat rises to the levels due to being lighter and cold air replaces it around the plant. Advection frosts occur when cold air blows into an area to replace warmer air that was present before the weather change. This can happen due to the displacement of cold air masses. Unlike the previous type, this frostbite can have a round-the-clock trend and be wider. Dealing with this type of frost is more difficult [3].

The characteristics of each kind of frosts shown in Table I. This study focuses on radiation frost prediction. Due to the fact that the forecast is made on each weather station, it is not possible to predict the arrival of cold air masses or sudden weather changes. Therefore, according to the temperature changes of each station during the day, it is possible to predict the occurrence of frost at night or the next day. In addition to this, we have tried to maximize the forecast time to take the necessary precautions.

There can be use two form of protection in frost occurrence. Passive methods are those that act in preventive terms, normally for a long period of time and whose action becomes particularly beneficial when freezing conditions occur. Active methods are temporary and they are energy or labor intensive, or both. active protection includes heaters, sprinklers and wind machines, which are used during the frost night to replace natural energy loss [3]. This study is effective in active prevention methods. Because by predicting the occurrence of frost, farmers can take the necessary measures to prevent damages. Also, in this case study, due to the semi-industrial nature of agriculture in Fars province, farmers in large scale need to know about the occurrence of frost. Our methods are often suitable for the active prevention model.

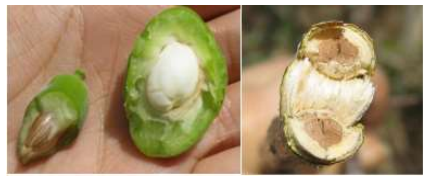

Fig. 1. Example of the occurrence of frost in pomegranate and chaghaleh plants [4]

A sequence of data collected in a timespan forms a time series. These data reflect the changes seen during a certain measured time. Therefore, it can be considered a time-dependent vector. As its name suggests, time series data refers to a set of data collected in specific time intervals. In this case, if x is a random process, the time series can be shown as the equations 1, where t represents time and it can also be a random variable.

$$x(t_0), x(t_1), x(t_2), ..., x(t_T), \quad (1)$$

A time series is basically classified as a dynamic data because the values of its features change with time, in other words, the value of each point in a time series consists of one or more observations that are measured at a specific moment.

Machine learning algorithms build a model based on sample data, known as training data, in order to make predictions or decisions without being explicitly programmed to do so. Deep learning is a branch of machine learning that base on neural networks which superimpose several layers to progressively learn higher-level features from the raw input. The most well-known example of deep learning is deep neural networks, which have been effective in many fields, including voice recognition, image classification, object recognition, and natural language processing. Deep learning is part of a broader family of machine learning algorithms used to learn data representations [5]. Gradient boosting is a machine learning technique used in regression and classification tasks, among others. It gives a prediction model in the form of an ensemble of weak learners, which are typically decision trees. When a decision tree is the weak learner, the resulting algorithm is called gradient-boosted trees and it usually outperforms random forest. A gradient-boosted trees model is built in a stage-wise fashion as in other boosting methods, but it generalizes the other methods by allowing optimization of an arbitrary differentiable loss function [6].

According to the contents stated in this section, the problem of forecasting frost is one of the important and prominent issues in the field of meteorology and agriculture, which causes a lot of damage to the Iran and Fars province every year, and also machine learning methods can be considered as an efficient method to solve different problems. For this purpose, in this study, we seek to provide an efficient method in terms of accuracy and time for the problem of frost prediction - or minimum temperature prediction at every day - using different methods, including deep learning and gradient boosting.

## II. REALTED WORK

Forecasting time series in each application has its own methods. To predict the time series of daily temperature - as it was said, the prediction of frost is actually the prediction of the time series of the minimum temperature - there are different approaches, each of which is mentioned below.

*A. Empirical method*

These methods are the oldest temperature forecasting methods and use the physical relationships that govern the atmosphere that are obtained experimentally, or model the minimum temperature through many observations and usually using linear regression. One of the most important methods is the method proposed by FAO. Most of the researches that have been done to improve forecasting have tried to compare the error rate of the proposed model with the FAO experimental model [3].

For example, in 2013, Qarakhani et al. investigated Linaker's experimental model for temperature modeling at least in four synoptic stations of Fars province and concluded that the experimental model of Linaker for Darudzen and Shiraz synoptic stations with RMSE of 1.66 and 1.93 respectively. It is considered a suitable method [7]. Linaker's empirical model is discussed in the results section.

In an article in 2004, Mohammad Nia Qaraei et al. presented an empirical formula for each of the synoptic stations of North Khorasan, Razavi and South Khorasan Provinces in the General Meteorology Department of Razavi Khorasan Province in 2004, which is used for the minimum temperature at night, which is exactly used for Frost prediction. With 12-year statistics from 1992 to 2003, they obtained an experimental model whose RMSE error was about 2 to 3 in different cities of this province [8].

*B. Statistical method*

Also, in 2008, Heydari Bani et al presented a statistical model in a case study for Shahrekord station, which is used to predict the limit temperatures in this station. Also, in another study, statistical modeling method and synoptic method were performed on Shahrekord station. They found that although it has a larger RMSE error of about one degree Celsius, but because it is a general model, it can be useful as an auxiliary and complementary tool in predicting threshold temperatures. They also suggested that the combination of synoptic-statistical models could possibly achieve better results [9].

In an article, Bazrafshan presented a hierarchical algorithm for the detection of radiative and advection glaciers by analyzing the meteorological data of 51 meteorological stations of the country. According to two temperature thresholds of 4 and 0 degrees Celsius, this algorithm separates the three groups of radiant, drifting and turbulent frosts for times of the year when frost occurred. They also drew a frost zoning map across the country [10].

*C. Artificial Intelligence method*

Qarakhani et al investigated the problem of minimum temperature modeling in Fars province for four stations by examining the one-layer artificial neural network model and concluded that the artificial neural network model works a little better than the experimental model. Their results using the neural network for the synoptic stations of Darudzen and Shiraz with RMSE were 1.58 and 1.75, respectively, which are better results than Linaker's experimental method [7].

In a study by Esfandiari et al. in 2013, they used the MLP multilayer perceptron model for the city of Saqqez, which obtained good results and reported the error of this method to be 0.8 siliceous at most. Also, in 2014, they changed their method to ANN model for prediction, and reported the same results for Saqez city [11].

In 2016, Zaytar et al presented a deep neural network architecture for use in meteorological data time series, which was based on LSTM stack networks. And it predicted the same sequence of meteorological values in two different 24-hour and 72-hour models for 9 cities in Morocco. The results showed that neural networks based on LSTM performed well compared to traditional methods and can be used as a general method for forecasting weather conditions [12].

In 2019, in a research conducted by Barooni et al., neural network methods such as multi-layer perceptron network and support vector machine were compared with empirical method. The output of this article was that the use of SVM method with RBF radial base kernel has less error compared to the other two models. In this research, they used the time series information of 4 meteorological stations [13]. Also, once again in that year, in another study, they were able to obtain more useful results in terms of RMSE error and model prediction time. With the use of LSTM neural network regression models and its modeling on 8 stations [14]. Due to the same case study in this article as well as the same data received from the same source, our research is a comparative research between new methods and these two articles.

## III. METHODS

As mentioned, three methods have been used in this article, which are described below in order of quality. All the methods run on Intel ® Xeon(R) CPU E5-269 2.20GHz.

### A. Gated Recurrent Unit (GRU)

GRU or Gated Recurrent Unit architecture was introduced in 2014 by Cho et al. This architecture is presented in order to solve the shortcomings of the traditional recurrent neural network, such as the vanishing gradient problem, as well as to reduce the overhead in the LSTM architecture. GRU is generally considered as a modified version of LSTM because both these architectures use the same design and in some cases equally achieve excellent results. We mentioned that to solve the vanishing gradient problem in the traditional neural network, one of the solutions is to use GRU. This type of architecture uses concepts called update gate and reset gate. These two so-called gates are basically two vectors that are used to decide what information is transmitted to the output and what information is not transmitted. The special thing about these gates is that these gates can be trained to retain information from much earlier steps without changing over time (between different time steps) [15]. In Fig. 2, it is shown the different between LSTM and GRU architecture.

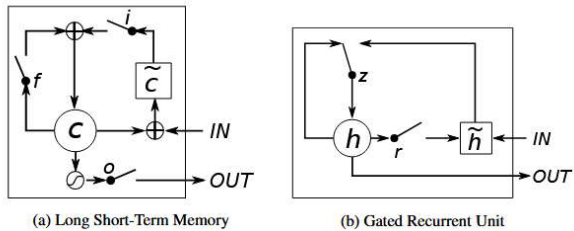

Fig. 2. Illustrate of (a) LSTM and (b) GRU with r and z are the reset and update gates [15]

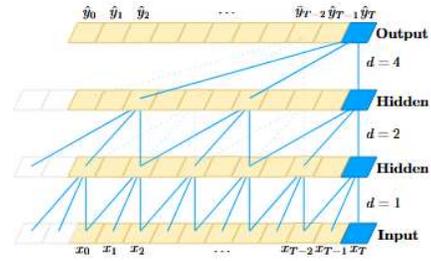

Fig. 3. Illustrate of (a) LSTM and (b) GRU with r and z are the reset and update gates [17]

### B. Temporal Convolutional Networks (TCN)

As a basic descriptive term for a family of architectures, that refers to the presented architecture as a temporal convolution network [16]. This architecture has important features. First, the convolutions in the architecture are causal, meaning that there is no leakage of information from the future to the past. Considering that we do not have any information about the future, this feature of TCNs is necessary for predicting the future temperature. Second, the architecture can take a sequence of any length and map it to an output sequence of the same length, just like an RNN. Considering that in the GRU network we had a series prediction, this case can be a good comparison between these two networks [17].

As it is clear in the Fig. 3, the architecture of TCN models is in a way that does not look into the future and also tries to consider wider receptive field. This means we can look at the input data with a wider viewing window. As mentioned, in meteorological data, the model have to be a causal, which TCN networks have this characteristic.

For both TCN and GRU methods, the same customized loss function is used according to the following equation.

$$Loss = (1/48 \times (y_{pred} - y_{true})^2 ) + |Min(y_{pred}) - Min(y_{true})| \quad (2)$$

The first part of equation 2 is Mean Square Error (MSE). But in the second part, we have the minimum temperature difference in the predicted and real data. This will optimize the neural network model to reduce this gap. It is important because we generally want to predict the minimum temperature. So instead of minimize the error for all future temperature sequence, we add a second term to try to optimize the difference between true and predicted minimum in the next day. This means that in practice we want to bring the prediction chart closer to the actual chart in troughs. This change will lead to beneficial improvements which we have discussed in the next part.

### C. Gradient Boosting (XGBoost)

Gradient boosting is a machine learning method for regression and classification problems that creates a predictive model in the form of a set of weak learners. Instead of training all the models separately, the "Boosting" process trains the models one after the other. Each new model is trained with the aim of correcting the errors caused by previous models. Models are added sequentially until there is no further improvement possible. The advantage of this iterative method is that the added models seek to correct the mistakes made by other models. In the standard ensemble classification method where models are trained individually, all models may make the same mistakes. Gradient boosting refers to a method in which new models are trained with the aim of predicting the residuals of previous models. The Fig. 4 shows how the gradient boosting algorithm works on regression problems. XGBoost is an algorithm that is recently used in the field of machine learning. The XGBoost algorithm is an implementation of decision tree gradient boosting designed for high speed and efficiency [18].

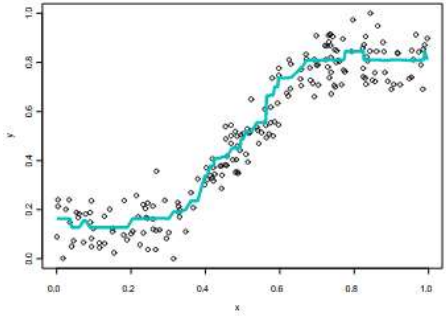

Fig. 4. Boosted regression on linear model [19]

It should be noted that the XGBOOST library uses different hyperparameters to fit the model better. The parameters in the model implemented are max-depth=3, n-estimators=200, colsample-bytree=0.5, booster=dart and rate-drop=0.1. These hyperparameters were tuned in the implementation and the best results were obtained using these parameters. These results are described in the next section.

For implementation, we first received the data of the desired stations. Then we used the three characteristics of these stations for prediction. These characteristics include minimum temperature, maximum temperature and dew point. The characteristics have been measured at each station as time series with half-hour intervals. So we have 48 feature set for each day. Our output in all three models is the next day's minimum temperature.

## IV. RESULTS

According to the models mentioned in the previous section the results obtained for each of the models are detailed in the respective tables.

TABLE II. GATED RECURRENT NETWORK (GRU)

| Stations | Kamfiroz | Korball | Eej | Jaahrom | Mamasani | Aliabad | Lar | Bavanat |
|---|---|---|---|---|---|---|---|---|
| Avg. Train RMSE | 1.52 | 1.65 | 2.05 | 1.68 | 1.76 | 1.56 | 1.61 | 1.96 |
| Best Train RMSE | 1.26 | 1.42 | 1.56 | 1.52 | 1.41 | 1.37 | 1.35 | 1.72 |
| Avg. Test RMSE | **2.26** | **1.99** | **2.02** | **1.97** | **1.87** | **1.86** | **1.79** | **2.49** |
| Best Test RMSE | 1.85 | 1.66 | 1.64 | 1.78 | 1.69 | 1.68 | 1.47 | 2.24 |

TABLE III. TEMPORAL CONVOLUTIONAL NETWORK (TCN)

| Stations | Kamfiroz | Korball | Eej | Jaahrom | Mamasani | Aliabad | Lar | Bavanat |
|---|---|---|---|---|---|---|---|---|
| Avg. Train RMSE | 1.29 | 1.32 | 1.42 | 1.49 | 1.43 | 1.33 | 1.26 | 1.76 |
| Best Train RMSE | 1.51 | 1.75 | 1.70 | 1.59 | 1.62 | 1.65 | 1.21 | 2.21 |
| Avg. Test RMSE | **1.80** | **2.04** | **2.07** | **1.74** | **1.87** | **1.80** | **1.33** | **2.36** |
| Best Test RMSE | 1.17 | 1.18 | 1.3 | 1.38 | 1.28 | 1.20 | 1.17 | 1.52 |

It should be mentioned that the RMSE error in the model means that the model was able to predict tomorrow's minimum temperature with this difference in degrees Celsius. This means that as thTABLE IV. is value gets closer to zero, the number that the model predicted for tomorrow's minimum temperature is closer to what recorded.

The table II is related to the GRU model. As can be seen, the specified errors are given for 8 meteorological stations. These errors include the mean RMSE error for the training and test data. Also, the best RMSE result in all runs is also Gated recurrent network (GRU) shown in the table II.

Also, the results obtained with the causal model of TCN are shown in the table III. As can be seen, deep learning models have a good performance on frost prediction. So that both their average RMSE error and the best result obtained in different epochs perform better than the practical method discussed below.

The results for XGBoost shown in Table IV. A remarkable point for the method implemented with XGBoost is that the difference between the best observed error and their mean is very small. This means that this method was able to achieve consistent results even with the addition of a random value in the modeling. This case can be investigated in the future. It is also clear that the RMSE of the test data in this method has improved compared to other methods. The difference between these methods examined in another table.

TABLE V. GRADIENT BOOSTING (XGBOOST)

| Stations | Kamfiroz | Korball | Eej | Jaahrom | Mamasani | Aliabad | Lar | Bavanat |
|---|---|---|---|---|---|---|---|---|
| Avg. Train RMSE | 1.10 | 1.16 | 1.03 | 1.31 | 0.98 | 1.26 | 1.00 | 1.73 |
| Best Train RMSE | 1.08 | 1.14 | 1.00 | 1.29 | 0.96 | 1.23 | 0.98 | 1.70 |
| Avg. Test RMSE | **1.60** | **1.57** | **1.48** | **1.60** | **1.47** | **1.53** | **1.22** | **2.04** |
| Best Test RMSE | 1.57 | 1.53 | 1.46 | 1.58 | 1.44 | 1.53 | 1.20 | 2.03 |

Using the proposed loss function can have positive effects. The horizontal axis shows the input and output. In this way, the first 48 units represent the first day as input and the second 48 units represent the series of the second day as output, which we want to predict the minimum temperature of. In this image, it is clear that if we use the proposed loss function instead of MSE, the model will reduce its distance to the minimum temperature. This allows us to provide a more accurate prediction in our problem. This result shown in Fig. 5.

The results obtained from the experimental model provided by FAO as well as the error difference on the test data in each of the three methods with the experimental results are given in the table V. As it is clear from this table, we can say that XGBoost method has given the best overall result in all meteorological stations. Also, the TCN method gets better results than the GRU method except for two stations, with a small difference, which can be due to its architectural properties. In only one station, we have seen the superiority of the experimental method over the GRU method. In addition to being very small, this difference is not very important considering the other results. Another important point is that the XGBoost method is far better thant other impelemented methods.

TABLE VI. COMPARE ALL METHODS WITH EMPRICAL

| Methods | Empirical | Differences with the empirical method | | |
|---|---|---|---|---|
| | | GRU | TCN | XGBoost |
| Kamfiroz | 1.91 | -0.35 | 0.11 | **0.31** |
| Korball | 2.20 | 0.21 | 0.16 | **0.63** |
| Eej | 2.22 | 0.20 | 0.15 | **0.74** |
| Jahrom | 3.56 | 1.59 | 1.82 | **1.96** |
| Mamasani | 3.84 | 1.97 | 1.97 | **2.37** |
| Aliabad | 2.31 | 0.45 | 0.51 | **0.78** |
| Lar | 6.43 | 4.64 | 5.10 | **5.21** |
| Bavanat | 2.74 | 0.25 | 0.38 | **0.70** |

## V. CONCLUSION

The aim of this study was to reduce frost damage by using timely prediction of its occurrence. Because the prediction time with current methods is only a few hours, which is not enough for preventive measures.

To solve the problem of frostbite, it is necessary to know its types and understand their characteristics. Considering that frostbite is one of the most important and harmful hazards in meteorology and agriculture, it is possible to prevent these damages through computer calculations and artificial intelligence modeling. In the modeling, we used three characteristics of minimum temperature, maximum temperature and dew point, which were sampled in 30 minute intervals. These data can be considered as a time series.

Sequence-by-sequence deep learning methods, in addition to being able to improve the RMSE error, can also provide more time to deal with freezing. As mentioned, these methods use sequence-by-sequence modeling to predict the minimum temperature of the next day that is part of this sequence. TCN and GRU methods are among these methods that TCN method provides better results in most stations. In these methods, by optimizing the loss function for these types of networks, an optimal weights was obtained. It has the ability to increase the prediction time and improve accuracy to a reasonable extent by using the nonlinear power characteristics of neural networks and time series property of data.

The gradient boosting method implemented by XGBoost, with the same input as deep learning models, can predict the minimum temperature with faster speed and less error, in addition to keeping the prediction time. This error is much better on average compare to other impelemented methods. In general, to solve the problem of predicting the minimum temperature, which is widely used in frostbite, the XGBoost method can offer us the best performance.

For future works, other features can be presented to the model. For example, solar radiation or wind speed, etc. It is also possible to use the information of the nearby stations of each station as input data, which can have a greater impact in predicting other types of frost.

## VI. ACKNOWLEDGEMENT

We are grateful to the Department of "New Technologies of Fars Province's Agricultural Jihad" and "General

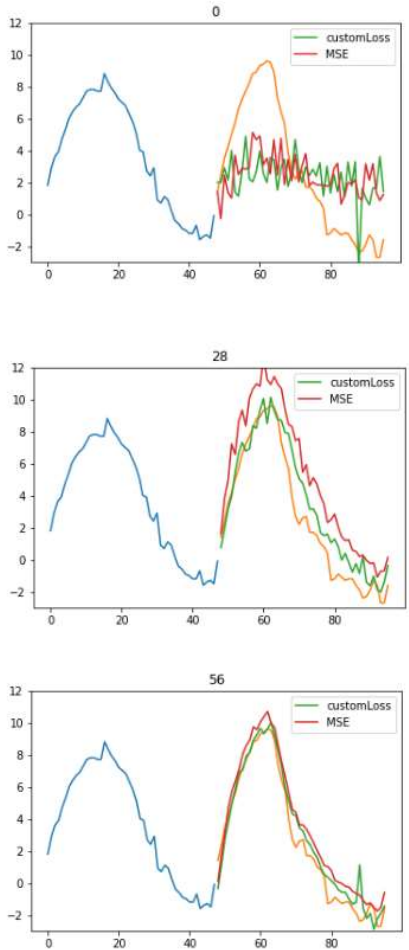

Fig. 5. Effectivness of custom loss function in different iteratinos according to MSE

Department of Meteorology of Fars Province" for providing the statistical data of the past years. Certainly, without the help of these orginizations, this research would not have been possible.## VII. References

[1] M. J. Nazemosadat, A. R. Sepaskhah, and S. Mohammadi, "A case study on the relationship between daily dewpoint and minimum temperature in next day in Jahrom in Iran," *Iranian Journal of Agricultural Science and Technology*, vol. 5, no. 3, pp. 9–17, 2001.

[2] A. Khalili, "Quantitative evaluation of spring frost risk to agricultural and horticultural crops in iranand modeling," Journal of Agricultural Meteorology, vol. 2, no. 1, 2014.

[3] H. G. Jones, "Frost protection: fundamentals, practice, and economics. Volume 1. By R. L. Snyder and J. P. de Melo-Abreu. Rome: FAO (2005), pp. 223, US38.00. ISBN 92-5-105328-6 Volume 2. By R. L. Snyder, J. P. de Melo-Abreu and S. Matulich. Rome: FAO (2005), pp. 64. US24.00. ISBN 92-5-10539-4," *Exp. Agric.*, vol. 42, no. 3, pp. 369–370, 2006.

[4] A. Asadi and S. Karbalaei, "A look at the damages of cold stress on agricultural products of the country in 2010," 2010.

[5] I. Goodfellow, Y. Bengio, and A. Courville, *"Deep Learning"*, London, England: MIT Press, 2016.

[6] J. H. Friedman, "Greedy function approximation: A gradient boosting machine," *Ann. Stat.*, vol. 29, no. 5, pp. 1189–1232, 2001.

[7] A. Gharehkhani, N. Ghahreman, and B. Bakhtiari, "Prediction of minimum temperature (chilling) using empirical models and artificial intelligence (Case study: Fars province, Iran)," The 2nd International conference of Plant, Water, Soil and Weather Modeling, Kerman, Iran, 2013.

[8] S. M. Qaraei, G. A. Rastgo, and S. Malbosi, "Presenting an experimental formula for local prediction of the minimum night temperature for each of the synoptic stations of North, Razavi and South Khorasan provinces,", Applied scientific conference on dealing with frostbite, Yazd, Iran, 2005.

[9] M. H. Beni, A. Samanipoor, A. Barati, and M. Shiasi, "Comparing Between Statistical And Synoptically Model In Forecasting Of Extreme Temperatures. (case study shahrekord synoptic station),", The First International conference on Plant, Water, Soil and Weather Modeling, Kerman, Iran, 2010.

[10] J. Bazrafshan, "Radiation, advection and mixed freezing and frost risk assessment and zoning in Iran," *Journal of Agricultural Meteorology*, vol. 2, no. 1, 2014.

[11] F. Esfandyari, S. A. Hosaini, H. Ahmadi, and K. Mohammadpour, "Predictive Modeling of Saghez Township Late Colds spring through Multilayer Perceptron (MLP) model,", The 2nd International conference of Plant, Water, Soil and Weather Modeling, Kerman, Iran, 2013.

[12] M. Akram and C. El, "Sequence to sequence weather forecasting with long short-term memory recurrent neural networks," *Int. J. Comput. Appl.*, vol. 143, no. 11, pp. 7–11, 2016.

[13] A. Barooni and K. Ziarati, "Modeling the minimum temperature to predict frost in Fars province using neural network, support vector machine and emprical models,", 4th International Congress of Developing Agriculture, Natural Resource, Environment and Tourism of Iran, Tabriz, Iran, 2019.

[14] A. Barooni and K. Ziarati, "Modeling minimum temperature in Fars province using LSTM recurrent neural network model,", 1st National Conference on Fundamental Researches in Agricultural and Environmental Science, Tehran, Iran, 2019.

[15] J. Chung, C. Gulcehre, K. Cho, and Y. Bengio, "Empirical evaluation of gated recurrent neural networks on sequence modeling," *arXiv [cs.NE]*, 2014.

[16] Lea, Colin, Michael D. Flynn, Rene Vidal, Austin Reiter, and Gregory D. Hager. "Temporal convolutional networks for action segmentation and detection." In *proceedings of the IEEE Conference on Computer Vision and Pattern Recognition*, pp. 156-165. 2017.

[17] Bai, Shaojie, J. Zico Kolter, and Vladlen Koltun. "An empirical evaluation of generic convolutional and recurrent networks for sequence modeling." *arXiv preprint arXiv:1803.01271* (2018).

[18] Chen, Tianqi, and Carlos Guestrin. "Xgboost: A scalable tree boosting system." In *Proceedings of the 22nd acm sigkdd international conference on knowledge discovery and data mining*, pp. 785-794. 2016.

[19] Ridgeway, Greg, David Madigan, and Thomas S. Richardson. "Boosting methodology for regression problems." In *Seventh International Workshop on Artificial Intelligence and Statistics*. PMLR, 1999.